%% file: acl_latex.tex
\documentclass[11pt]{article}

\usepackage[final]{acl}

\usepackage{times}
\usepackage{latexsym}

\usepackage[T1]{fontenc}

\usepackage[utf8]{inputenc}

\usepackage{microtype}

\usepackage{inconsolata}

\usepackage{graphicx}

\usepackage{adjustbox}
\usepackage{booktabs}
\usepackage{enumitem}
\usepackage[table]{xcolor}

\input{math_commands}

\newcommand{\numtoroman}[1]{\romannumeral #1}

\title{Optimizing User Profiles via Contextual Bandits for Retrieval-Augmented LLM Personalization}

\author{
    \bfseries
    Linfeng Du$^{1,6}$ \quad
    Ye Yuan$^{1,6}$ \quad
    Zichen Zhao$^{2}$ \\
    \bfseries
    Fuyuan Lyu$^{1,6}$ \quad
    Emiliano Penaloza$^{3,6}$ \quad
    Xiuying Chen$^{2}$ \quad
    Zipeng Sun$^{1,6}$ \\
    \bfseries
    Jikun Kang$^{4}$ \quad
    Laurent Charlin$^{5,6}$ \quad
    Xue Liu$^{1,2,6}$ \quad
    Haolun Wu$^{1,6}$\setcounter{footnote}{3}\thanks{Corresponding author.} \\
    $^{1}$McGill University \quad
    $^{2}$Mohamed bin Zayed University of Artificial Intelligence \\
    $^{3}$Université de Montréal \quad
    $^{4}$Salesforce \quad
    $^{5}$HEC Montréal \quad
    $^{6}$Mila - Quebec AI Institute \\
    \texttt{\{linfeng.du, haolun.wu\}@mail.mcgill.ca} \\
}

\begin{document}
\maketitle

\input{sections/00-abstract}
\input{sections/01-introduction}
\input{sections/02-related}
\input{sections/03-methodology}
\input{sections/04-setup}
\input{sections/05-results}
\input{sections/06-conclusion}
\input{sections/07-limitations}

\bibliography{custom}

\input{sections/99-appendix}

\end{document}

%% file: math_commands.tex
\usepackage{amsmath,amsfonts,bm}

\def\eqref#1{equation~\ref{#1}}
\def\1{\bm{1}}

\def\vr{{\bm{r}}}

\DeclareMathAlphabet{\mathsfit}{\encodingdefault}{\sfdefault}{m}{sl}
\SetMathAlphabet{\mathsfit}{bold}{\encodingdefault}{\sfdefault}{bx}{n}

\newcommand{\E}{\mathbb{E}}

%% file: sections/00-abstract.tex
\begin{abstract}
    Large language models (LLMs) excel at general-purpose tasks, yet adapting their responses to individual users remains challenging.
    Retrieval augmentation provides a lightweight alternative to fine-tuning by conditioning LLMs on user history records, and existing approaches typically select these records based on semantic relevance.
    We argue that relevance serves as an unreliable proxy for utility: a record may be semantically similar to a query yet fail to improve generation quality or even degrade it due to redundancy or conflicting information.
    To bridge this gap, we propose \textbf{PURPLE}, a contextual bandit framework that o\underline{P}timizes \underline{U}se\underline{R} \underline{P}rofiles for \underline{L}LM p\underline{E}rsonalization.
    In contrast to a greedy selection of the most relevant records, PURPLE treats profile construction as an order-sensitive generation process and utilizes a Plackett-Luce ranking model to capture complex inter-record dependencies.
    By training with semantically rich feedback provided by the likelihood of the reference response, our method aligns retrieval directly with generation quality.
    Extensive experiments on nine personalization tasks demonstrate that PURPLE consistently outperforms strong heuristic and retrieval-augmented baselines in both effectiveness and efficiency, establishing a principled and scalable solution for optimizing user profiles\footnote{Our code is available \href{https://github.com/linfeng-du/PURPLE}{here}.}.
\end{abstract}

%% file: sections/01-introduction.tex
\begin{figure}[t]
    \centering
    \includegraphics[width=0.95\linewidth]{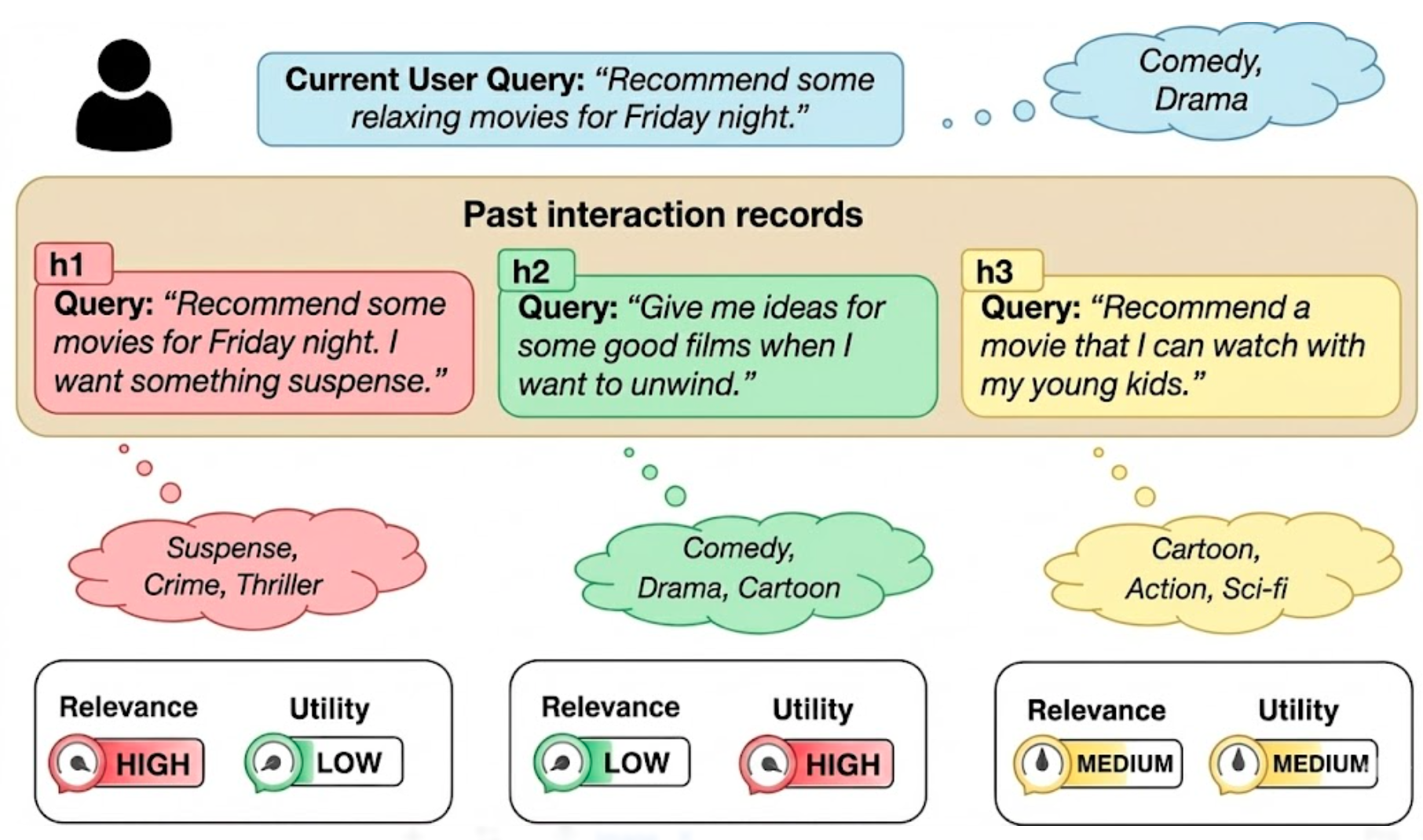}
    \caption{An illustration of the discrepancy between relevance and utility for a user's past interaction records given a query. The thought bubbles illustrate the potential movie genres that satisfy the underlying user intent.}
    \vspace{-4mm}
    \label{fig:example}
\end{figure}

\section{Introduction}
\label{sec:introduction}

Large language models (LLMs) have demonstrated remarkable success in various natural language processing tasks.
As these models are increasingly applied to personalized applications, tailoring responses to individuals based on their own preferences has become a crucial challenge.
Existing approaches for personalizing LLMs, such as reinforcement learning from human feedback (RLHF)~\citep{ouyang2022training}, generally require modifying model parameters.
These approaches incur high computational costs, demand frequent updates, and are impractical for real-time personalization at scale, especially when the LLM is not fully open-sourced.
Moreover, continually fine-tuning models for different individuals would complicate safety evaluation since each personalized variant would require separate testing.

In this paper, we study a lightweight approach to LLM personalization via retrieval augmentation~\citep{wu2025retrievalaugmented}, in which user profiles are constructed by retrieving and injecting past user records into the prompt.
Prior work shows that such profiles can effectively steer LLM outputs toward individual preferences~\citep{salemi2024lamp,jiang2025know}.
Compared to parameter-updating methods, retrieval-augmented personalization is lightweight, transparent, and readily deployable, as users can directly inspect and modify the records guiding generation.
However, naively including the full user history introduces redundancy and noise, and may exceed the model’s context window, while aggressive pruning risks discarding critical personalization signals.
This tension raises a central challenge: \textbf{\textit{which} user records should be selected to construct a user profile that maximizes personalization utility}, i.e., downstream task performance when conditioned on by the LLM?

Existing strategies for building user profiles often rely on heuristics, such as selecting records with the highest \textit{relevance}, defined as lexical or semantic similarity to the query~\citep{karpukhin2020dense}.
However, we argue that \textit{relevance} is not a reliable proxy for \textit{utility}, which measures the extent to which a record helps the LLM generate an appropriate personalized response.
Relevance-based heuristics conflate surface-level contextual similarity with the underlying preference signals that drive personalization, prioritizing records that look similar to the query rather than those that support the user's current intent.
To illustrate this dichotomy, consider Figure~\ref{fig:example}, where the user seeks \textit{``relaxing movies for Friday night''}.
Relevance-based heuristics prioritize record $h_1$ due to the high lexical overlap of \textit{``Friday night''}, ignoring the latent preference for \textit{``suspense''} which contradicts the current need for relaxation.
Conversely, record $h_2$ offers high utility by capturing the semantic intent to \textit{``unwind''}, despite lacking surface-level keyword matches.
This highlights that relevance scores prioritize how similar a record looks to the query, whereas utility depends on whether it supports the underlying user intent.

Beyond evaluating individual records, constructing an effective user profile requires list-aware reasoning rather than independent selection~\citep{ai2018learning, pang2020setrank}.
Even if the utility of each record could be estimated accurately in isolation, greedy top-$k$ selection remains suboptimal because record utilities do not compose additively.
Instead, the value of a record is contingent on the context provided by others, as irrelevant or conflicting information within the retrieved profile can significantly degrade downstream generation~\citep{liu2024lost, yoran2024making}.
Revisiting Figure~\ref{fig:example}, while $h_2$ and $h_3$ individually appear useful by reflecting a shared preference for lighthearted content, their combination is suboptimal.
Jointly prompting with $h_2$ and $h_3$ introduces additional constraints by combining \textit{``comedy''} preferences with \textit{``young kids''} requirements, which could dilute the conditioning signal, thereby degrading generation quality relative to using $h_2$ alone.
This confirms that personalization utility is non-monotonic.
Simply accumulating \textit{useful} records may introduce interference and actively degrade performance.

The limitations highlighted above call for a reranking mechanism that is directly optimized for downstream generation and sensitive to interactions among records.
Existing approaches fall short of these requirements: heuristic retrieval relies on static proxies like similarity, while recent listwise rerankers, though capable of modeling dependencies, remain constrained by relevance-oriented supervision.
To bridge this gap, we propose \textbf{PURPLE}, a framework that o\underline{P}timizes \underline{U}se\underline{R} \underline{P}rofiles for \underline{L}LM p\underline{E}rsonalization via contextual bandits~\citep{langford2007epochgreedy}.
In this formulation, the context consists of both the current query and the user's past records.
The selection policy is guided by a reward function reflecting downstream personalized generation performance.
PURPLE outputs a \textit{propensity score} for each user record, which parameterizes a Plackett-Luce ranking model to produce the final user profile.
This formulation enables the model to capture interactions between records and adaptively select those that are most beneficial as a whole.
We train PURPLE end-to-end using the policy gradient method~\citep{sutton1999policy}.

Our main contributions are as follows:
\begin{itemize}[left=0pt]
    \item We introduce PURPLE, a framework that casts retrieval-augmented LLM personalization as a contextual bandit problem, adaptively optimizing user profiles beyond static heuristics.
    \item We show through extensive experiments on nine personalization tasks, covering classification, regression, and short- and long-text generation, that PURPLE consistently outperforms strong baselines in both effectiveness and efficiency.
    \item We perform comprehensive ablation studies on PURPLE’s key design choices and further validate its effectiveness through extended analysis and human evaluation.
\end{itemize}

%% file: sections/02-related.tex
\section{Related work}
\label{sec:related}

\paragraph{LLMs for Personalization.}

LLMs demonstrate strong performance across domains~\citep{openai2024gpt4}, yet their outputs often diverge from user expectations because pre-training captures general rather than individual needs.
Reinforcement learning from human feedback (RLHF)~\citep{ouyang2022training} is widely employed to align pre-trained models with user preferences; however, while parameter-efficient fine-tuning (PEFT) (e.g., LoRA~\citep{hu2022lora}) drastically lowers its training costs, the necessity of model fine-tuning remains an impractical barrier for individual end users.
A complementary direction personalizes LLMs through user profiles~\citep{salemi2024lamp} constructed from prior user interactions.
Integrating these profiles has demonstrated substantial benefits across various tasks requiring personalization, including recommendation~\citep{penaloza2025tears}, summarization~\citep{zhang2024personalsum}, question answering~\citep{wu2024understanding}, content generation~\citep{shen2024pmg}, and chatbot interaction~\citep{jiang2025know}.
Yet, it remains unclear which specific user records within a profile truly drive performance improvements, particularly in retrieval-augmented generation (RAG), where efficacy hinges on retrieving semantically relevant context.
Moreover, little analysis has been conducted on how to optimally select and assemble user records into profiles with high personalization utility.
Our work addresses this gap by investigating how user profiles shape personalization in retrieval-augmented LLMs, and by proposing strategies for selecting user records to maximize downstream performance.

\paragraph{Retrieval-Augmented Generation.}

Retrieval-augmented generation (RAG) augments LLMs with external knowledge sources to enhance factuality and coverage.
Early architectures, such as REALM~\citep{guu2020retrieval} and RAG~\citep{lewis2020retrievalaugmented}, jointly optimize the retriever and the LM, while Re2G~\citep{glass2022re2g} further incorporates a reranking module trained end-to-end with the other components.
To mitigate computational costs, subsequent methods freeze the LM and perform in-context retrieval.
For instance, IC-RALM~\citep{ram2023incontext} leverages LLMs for reranking, whereas REPLUG~\citep{shi2024replug} distills retrievers from LLMs.
More recently, instruction-tuned variants like SelfRAG~\citep{asai2024selfrag} and RankRAG~\citep{yu2024rankrag} jointly model retrieval and generation, but their reliance on large-scale fine-tuning renders them impractical for user-level personalization.

The methods most relevant to our work are IC-RALM and REPLUG; however, both methods process retrieved records in isolation, which is a limitation our method directly addresses.
Specifically, REPLUG approximates joint reasoning by weighting individual generation outputs with retrieval probabilities, while IC-RALM periodically triggers retrieval during decoding and replaces previously used context.
These designs arise because jointly reasoning over multiple records leads to a combinatorial explosion in the number of possible profiles.
In contrast, our approach overcomes this limitation by modeling inter-record dependencies and directly optimizing over multi-record profiles without resorting to such approximations.

\paragraph{LLMs for Reranking.}

LLM-based rerankers are broadly categorized into pointwise, pairwise, and listwise paradigms.
Pointwise models, such as MonoBERT~\citep{nogueira2019multistage} and MonoT5~\citep{nogueira2020document}, score each query--document pair independently, whereas pairwise models like DuoT5~\citep{pradeep2021expandomonoduo} evaluate candidates relative to one another.
In contrast, listwise approaches such as ListT5~\citep{yoon2024listt5} and EBCAR~\citep{yuan2026embeddingbased} jointly model the full candidate set.
As LLMs continue to expand their context lengths, the listwise paradigm has seen rapid advancement through prompt-only ranking (RankGPT~\citep{sun2023chatgpt}), model distillation (e.g., RankVicuna, RankZephyr, Lit5Distill, FIRST~\citep{pradeep2023rankvicuna,pradeep2023rankzephyr,tamber2023scaling,gangireddy2024first}), and inference-time relevance extraction (ICR~\citep{chen2025attention}).
However, these methods inherently conflate relevance with utility, which is not ideal for personalization.
In this work, we instead train rerankers using downstream generation quality as a feedback signal, prioritizing task utility over semantic similarity.

%% file: sections/03-methodology.tex
\begin{figure*}
    \centering
    \includegraphics[width=\textwidth]{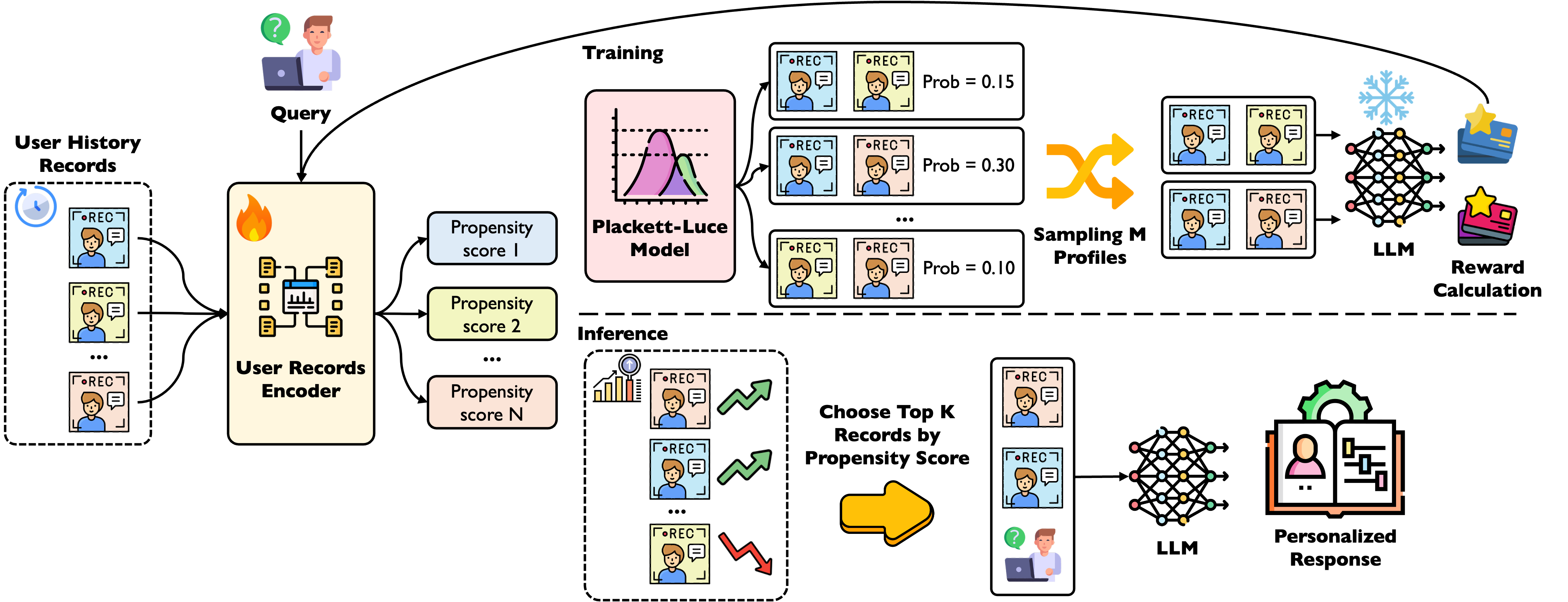}
    \caption{Workflow of the proposed PURPLE framework. User records encoder takes a user query and a list of user history records as input, outputting the propensity scores of all records. During \textbf{training}, a Plackett-Luce model is employed to convert the propensity scores into a probability distribution over all possible profiles, followed by sampling $M$ profiles for policy gradient estimation. During \textbf{inference}, records with the top-$K$ propensity scores are provided to the LLM along with the user query to generate a personalized response.}
    \label{fig:purple}
\end{figure*}

\section{Methodology}
\label{sec:methodology}

We formulate retrieval-augmented LLM personalization as a contextual bandit problem~\citep{langford2007epochgreedy}, where the goal is to learn a policy that selects informative user records.
Unlike classic multi-armed bandits, contextual bandits incorporate auxiliary information (e.g., the current query and user history) before making a selection, enabling direct optimization of retrieval strategies through policy gradient reinforcement learning.
This aligns the selection of user records with downstream personalization objectives.

\subsection{Problem Formulation}

We consider a dataset $\mathcal{D} = \{(\mathcal{H}^u, x^u, y^u)\}_{u=1}^{|\mathcal{D}|}$, where each example consists of a user’s collection of history records $\mathcal{H}^u$, a query $x^u$, and a reference personalized response $y^u$.
Personalization is achieved by retrieving informative records from $\mathcal{H}^u$ and supplying them as context to a frozen LLM, which then generates the final response.
In practice, we apply PURPLE as a reranking module on top of a candidate pool selected by lightweight heuristics, ensuring low-latency inference compatible with large-scale systems.
In the following development, we focus on a single user and omit the superscript index for brevity.

Let $\mathcal{H} = \{h_1, \dots, h_N\}$ denote the set of $N$ history records for a user, where each record $h_i = (x_i, y_i)$ is an input--output pair (e.g., a query and its answer from the user).
Given a new query $x$, our goal is to construct a user profile from $\mathcal{H}$ to condition the LLM for generating a personalized response.
Formally, a profile is an ordered tuple $\mathcal{P} = \langle p_1, \dots, p_K \rangle \in \mathrm{Perm}_K(\mathcal{H})$, which is a $K$-permutation of $\mathcal{H}$.
We stress that the profile is order-sensitive: different permutations of the same $K$ records correspond to distinct profiles and thus provide different inputs to the downstream LLM.

We formulate the selection of $\mathcal{P}$ as a \textit{contextual bandit problem}, where the context is given by the user’s history $\mathcal{H}$ and the query $x$, and the action corresponds to selecting $K$ records from $\mathcal{H}$ to construct a profile.
Formally, this formulation consists of the following key components:
\begin{itemize}
    \item \textbf{Context:} $\mathcal{C} = (\mathcal{H}, x)$, where $\mathcal{H}$ is the user's collection of history records and $x$ is the query.
    This representation captures both past user preferences and the immediate intent.
    \item \textbf{Actions:} $\mathcal{P} = \langle p_1, \dots, p_K \rangle \in \mathrm{Perm}_K(\mathcal{H})$, which corresponds to selecting $K$ distinct records from $\mathcal{H}$ in a particular order.
    The action thus determines not only which records to use but also how they are arranged.
    The size of the action space is $N! / (N-K)!$.
    \item \textbf{Reward:} $R(\mathrm{LLM}(\mathcal{P} \,\|\, x), y)$, a function that measures the quality of the LLM-generated response conditioned on the concatenation of the user profile $\mathcal{P}$ and the query $x$, relative to the reference personalized response $y$.
\end{itemize}

We model the policy with a neural distribution $\pi_{\theta}(\cdot \mid \mathcal{C})$, parameterized by $\theta$, which assigns probabilities to candidate user profiles given the context $\mathcal{C}$.
The objective is to learn parameters $\theta$ such that the policy assigns higher probabilities to more informative profiles, which ultimately enhance personalized text generation.
To this end, we maximize the expected reward over sampled user profiles, optimizing the following objective on a dataset $\mathcal{D}$ spanning multiple users, each associated with a set of history records, a query, and the corresponding reference response:
\begin{equation}
    \begin{aligned}
        \mathcal{J}&(\theta) = \\
        &\E_{(\mathcal{H}, x, y) \sim \mathcal{D}, \mathcal{P} \sim \pi_{\theta}(\cdot \mid \mathcal{C})}[R(\mathrm{LLM}(\mathcal{P} \,\|\, x), y)].
    \end{aligned}
    \label{eq:objective}
\end{equation}

It is challenging to directly optimize Equation~\ref{eq:objective} since the reward is not differentiable.
To address this, we employ the likelihood ratio gradient estimator from reinforcement learning and stochastic optimization~\citep{williams1992simple, sutton1999policy}, which allows us to compute the gradient as:
\begin{equation}
    \begin{aligned}
        \nabla_{\theta} \mathcal{J}&(\theta) = \E_{(\mathcal{H}, x, y) \sim \mathcal{D}, \mathcal{P} \sim \pi_{\theta}(\cdot \mid \mathcal{C})} \\
        & [\nabla_{\theta} \log\pi_{\theta}(\mathcal{P} \mid \mathcal{C}) R(\mathrm{LLM}(\mathcal{P} \,\|\, x), y)].
    \end{aligned}
    \label{eq:gradient}
\end{equation}
Since it is intractable to enumerate all profiles $\mathcal{P} \in \mathrm{Perm}_K(\mathcal{H})$ during the optimization process, we estimate Equation~\ref{eq:gradient} by randomly sampling $M = 32$ profiles.
To stabilize training and reduce variance in gradient estimation, we apply z-score normalization across the $M$ sampled rewards for each example.
The precise formulation of this estimator is detailed in Equation~\ref{eq:monte_carlo} of Appendix~\ref{app:gradient}.

\subsection{Model and Function Design}

\paragraph{Design of $\pi_{\theta}(\cdot \mid \mathcal{C})$}

Since different permutations of the selected records may lead to different responses, we adopt the Plackett–Luce (PL) ranking model, which assigns probabilities to profiles based on the scores of individual user records.
Therefore, $\pi_{\theta}(\cdot \mid \mathcal{C})$ defines a distribution over all $N! / (N - K)!$ permutations of length $K$ drawn from the user's $N$ history records.
The probability assigned to a specific profile $\mathcal{P}$ is given by:
\begin{equation}
    \pi_{\theta}(\mathcal{P} \mid \mathcal{C}) = \prod_{k=1}^{K} \frac{f_{\theta}(p_k; \mathcal{C})}{S - \sum_{j=1}^{k-1} f_{\theta}(p_j; \mathcal{C}))},
    \label{eq:plackett-luce}
\end{equation}
where $S = \sum_{i=1}^{N} f_{\theta}(h_i; \mathcal{C})$, and $f_{\theta}(\cdot)$ is the user record encoder that outputs a propensity score in $[0, 1]$ for each record, indicating the model’s tendency to include that record in the user profile.
During training, profiles are generated by sampling $K$ records without replacement based on Equation~\ref{eq:plackett-luce}.
At inference time, the top-$K$ records ranked by propensity scores are selected.
Because the PL model is order-sensitive and rewards are assigned to ordered sets, the learned propensity score can be interpreted as each record’s contribution to the selected profile.
We empirically demonstrate in Section~\ref{sec:topk} that this ordering achieves higher final utility compared with other baselines.

\paragraph{Design of $f_{\theta}$}

For the user record encoder $f_{\theta}$, we aim to capture the interdependencies among user records.
A key design consideration is the trade-off between modeling dependencies at the token level versus the record level.
While the former could, in principle, capture finer-grained interactions, it would quickly exceed the encoder’s context length.
To address this, we adopt a late interaction strategy~\citep{khattab2020colbert}, where we first obtain record embeddings with a pre-trained encoder, and then apply a Transformer encoder to model dependencies across records.
Figure~\ref{fig:purple} illustrates the overall workflow of our method.
We utilize a pre-trained Contriever~\citep{izacard2022unsupervised} to obtain token embeddings for both the query and the records.
Each record first cross-attends to the query at the token level, producing query-fused token representations that incorporate query information.
A subsequent pooling operation is applied to produce fixed-size record embeddings, which are then processed by a Transformer encoder to model inter-record dependencies.
We omit positional encodings to avoid ordering bias among records.

\paragraph{Design of Reward Function}

In this work, we propose an LLM-driven reward, where the policy is trained to maximize the log-likelihood that the LLM assigns to the reference response.
Formally, given a user profile $\mathcal{P}$, a query $x$, and a reference personalized response $y$, we define the reward as:
\begin{equation}
    \begin{aligned}
        R(\mathrm{LLM}&(\mathcal{P} \,\|\, x), y) = \log p_{\phi}(y \mid \mathcal{P}, x) \\
        &= \sum_{j=1}^{|y|} \log p_{\phi}( y_j \mid \mathcal{P}, x, y_{<j}),
    \end{aligned}
\end{equation}
where $\phi$ are the parameters of the LLM, and $p_{\phi}(\cdot)$ denotes its next-token distribution.
Unlike task-specific metrics such as accuracy or ROUGE-1, which only provide coarse-grained feedback on the final answer~\citep{liu2025nover}, the log-likelihood of the reference response serves as a semantically richer reward signal.
The increased granularity allows the policy to distinguish between plausible and optimal profiles even when the final metrics might be identical.
Empirical results in Section~\ref{sec:ablation} demonstrate the advantages of this reward over metric-based formulations.

Moreover, we show in Appendix~\ref{app:reward} that, with the log-likelihood reward, the objective defined in Equation~\ref{eq:objective} corresponds to maximizing the evidence lower bound (ELBO) of the marginalization-based RAG formulation~\citep{lewis2020retrievalaugmented}.
Directly optimizing this marginal likelihood is intractable in our setting due to the combinatorial space of possible profiles, whereas our objective provides a tractable surrogate with rich learning signals.
In the next section, we empirically demonstrate that this log-likelihood-based reward is robust across different types of downstream tasks.

%% file: sections/04-setup.tex
\section{Experimental Setup}
\label{sec:setup}

\input{tables/lamp}

\subsection{Dataset and Evaluation}

We evaluate the performance of PURPLE using \texttt{Phi-4-Mini-Instruct}~\citep{microsoft2025phi4mini} and \texttt{Llama-3-8B-Instruct}~\citep{grattafiori2024llama} as the frozen LLM, and further scale to \texttt{Llama-3-70B-Instruct}.
Our experiments span a range of personalization settings, including personalized classification, regression, and both short- and long-text generation from the LaMP~\citep{salemi2024lamp} and LongLaMP~\citep{kumar2024longlamp} benchmarks.
We follow the prompt templates of \citet{salemi2024lamp} and \citet{kumar2024longlamp} to incorporate user profiles into the original queries.

Specifically, we evaluate PURPLE on \textbf{nine personalization tasks}, including two classification tasks — \textit{Personalized Citation Identification} (Citation) and \textit{Personalized Movie Tagging} (Movie) — evaluated with Accuracy and F1; one regression task — \textit{Personalized Product Rating} (Rating) — evaluated with MAE and RMSE; and six generation tasks, evaluated with ROUGE-1 (RG1), ROUGE-L (RGL)~\citep{lin2004rouge}, and METEOR (MT)~\citep{banerjee2005meteor}.
The generation tasks are further divided into short-text generation — \textit{Personalized News Headline Generation} (News), \textit{Personalized Scholarly Title Generation} (Scholar), and \textit{Personalized Tweet Paraphrasing} (Tweet) — and long-text generation — \textit{Personalized Abstract Generation} (Abstract), \textit{Personalized Topic Generation} (Topic), and \textit{Personalized Product Review Generation} (Review).
In all experiments, we first use Contriever~\citep{izacard2022unsupervised} to retrieve 20 candidate records as the user history $\mathcal{H}$, and then select 5 of them to construct the user profile $\mathcal{P}$.
Appendix~\ref{app:implementation} further elaborates implementation details.

\subsection{Baseline Methods}

We focus on the setting where the LLM is kept frozen and no ground-truth profile is available for training the reranker.
Therefore, we compare against three categories of prior methods that likewise neither fine-tune the LLM nor rely on supervision from ground-truth profiles.
Methods that operate under different assumptions or settings are not included in our main comparisons, as these differences prevent a fair and controlled evaluation.
We discuss these methods in detail in Appendix~\ref{app:alternative}.

The baselines we compare with include \textbf{(\numtoroman{1}) Zero-Shot Rerankers}, represented by RankGPT~\cite{sun2023chatgpt} and ICR~\cite{chen2025attention}.
For both methods, we adopt \texttt{Llama-3-8B-Instruct} as the reranker LLM.
We also report the performance of RankGPT with \texttt{GPT-5-nano} to reflect an upper bound of the methods that distill from the ranking results of state-of-the-art proprietary LLMs~\citep{pradeep2023rankvicuna,pradeep2023rankzephyr,tamber2023scaling,gangireddy2024first}.
\textbf{{(\numtoroman{2})} In-Context Retrieval-Augmented Language Models (RALMs)}, represented by IC-RALM~\cite{ram2023incontext} and REPLUG~\cite{shi2024replug}.
Both methods consider each record in isolation when generating a response.
They incorporate multiple records from the user profile either through context switching (IC-RALM) or marginalization (REPLUG).
Additionally, we include \textbf{(\numtoroman{3}) Efficient Sparse and Dense Retrievers}, applied directly as rerankers.
We include BM25~\cite{robertson2009probabilistic} for the sparse retriever and Contriever~\cite{izacard2022unsupervised} for the dense retriever.
These methods represent the efficiency-oriented side of the efficiency--performance trade-off.

%% file: tables/lamp.tex
\begin{table*}[t]
    \centering
    \caption{Results on the LaMP benchmark. The best and second-best results in each column are highlighted in \textbf{bold} and \underline{underlined}, respectively. We report the mean and standard deviation over three runs with different random seeds for LLM generation. For \texttt{GPT-5-nano}, we report only a single run due to API cost constraints.}
    \adjustbox{max width=\linewidth}{
    \begin{tabular}{l|cc|c|ccc}
        \toprule
        \textbf{Task}
            & \textbf{Citation}
            & \textbf{Movie}
            & \textbf{Rating}
            & \textbf{News}
            & \textbf{Scholar}
            & \textbf{Tweet} \\
        \midrule
        \textbf{Metric}
            & Acc. / F1
            & Acc. / F1
            & MAE / RMSE
            & RG1 / RGL / MT
            & RG1 / RGL / MT
            & RG1 / RGL / MT \\
    \midrule
    \multicolumn{1}{l}{\textbf{\textit{With Phi-4-Mini-Instruct (3.84B)}}} & \multicolumn{6}{l}{} \\
    \midrule
        BM25
        & $63.9_{0.53}$ / $63.8_{0.69}$
        & $34.5_{0.65}$ / $29.6_{0.61}$
        & $0.438_{0.00}$ / $0.852_{0.01}$
        & $14.4_{0.16}$ / $12.8_{0.16}$ / $11.9_{0.12}$
        & $39.7_{0.08}$ / $33.2_{0.04}$ / $42.2_{0.08}$
        & $38.3_{0.04}$ / $33.6_{0.08}$ / $35.2_{0.04}$ \\
        Contriever
        & $64.6_{0.33}$ / $64.5_{0.20}$
        & $36.0_{0.37}$ / $31.1_{0.24}$
        & $\underline{0.424}_{0.01}$ / $\underline{0.830}_{0.03}$
        & $14.6_{0.04}$ / $13.1_{0.00}$ / $12.2_{0.04}$
        & $39.7_{0.00}$ / $33.3_{0.08}$ / $42.0_{0.08}$
        & $38.6_{0.08}$ / $33.7_{0.08}$ / $35.8_{0.04}$ \\
        IC-RALM
        & $62.0_{0.20}$ / $61.7_{0.37}$
        & $33.6_{0.12}$ / $28.8_{0.16}$
        & $0.471_{0.01}$ / $0.857_{0.02}$
        & $13.4_{0.04}$ / $11.9_{0.04}$ / $11.0_{0.00}$
        & $37.6_{0.12}$ / $31.0_{0.12}$ / $41.0_{0.33}$
        & $38.1_{0.12}$ / $33.3_{0.16}$ / $35.2_{0.12}$ \\
        REPLUG
        & $61.2_{0.94}$ / $60.2_{0.78}$
        & $\underline{37.4}_{0.49}$ / $\underline{32.1}_{0.37}$
        & $0.486_{0.01}$ / $0.891_{0.02}$
        & $14.1_{0.04}$ / $12.7_{0.00}$ / $11.4_{0.08}$
        & $38.1_{1.63}$ / $32.8_{0.90}$ / $40.4_{1.51}$
        & $\textbf{42.2}_{0.08}$ / $\textbf{37.3}_{0.00}$ / $\textbf{38.6}_{0.04}$ \\
        RankGPT (Llama-3-8B-Instruct)
        & $63.9_{0.82}$ / $63.7_{0.65}$
        & $34.2_{0.94}$ / $28.9_{1.18}$
        & $0.446_{0.00}$ / $0.863_{0.01}$
        & $14.7_{0.33}$ / $13.1_{0.24}$ / $12.3_{0.24}$
        & $39.7_{0.00}$ / $33.3_{0.04}$ / $42.0_{0.04}$
        & $38.2_{0.04}$ / $33.4_{0.02}$ / $35.2_{0.08}$ \\
        RankGPT (GPT-5-nano)
        & \underline{65.9} / \textbf{65.6}
        & 35.5 / 31.4
        & 0.444 / 0.865
        & 14.6 / 13.0 / 12.1
        & \underline{39.8} / \underline{33.4} / \underline{42.3}
        & 38.5 / 33.7 / 35.5 \\
        ICR (Llama-3-8B-Instruct)
        & $65.2_{0.53}$ / $\underline{65.0}_{0.45}$
        & $34.1_{0.73}$ / $29.8_{1.02}$
        & $\underline{0.424}_{0.00}$ / $\underline{0.830}_{0.02}$
        & $\underline{15.0}_{0.00}$ / $\underline{13.4}_{0.04}$ / $\underline{12.5}_{0.00}$
        & $39.5_{0.08}$ / $33.1_{0.08}$ / $42.0_{0.04}$
        & $38.6_{0.12}$ / $33.6_{0.24}$ / $35.5_{0.16}$ \\
        \rowcolor{green!15} PURPLE (Ours)
        & $\textbf{66.0}_{0.20}$ / $\textbf{65.6}_{0.16}$
        & $\textbf{38.6}_{0.33}$ / $\textbf{34.2}_{0.53}$
        & $\textbf{0.419}_{0.01}$ / $\textbf{0.808}_{0.02}$
        & $\textbf{15.1}_{0.04}$ / $\textbf{13.5}_{0.00}$ / $\textbf{12.6}_{0.08}$
        & $\textbf{40.0}_{0.04}$ / $\textbf{33.5}_{0.04}$ / $\textbf{42.4}_{0.00}$
        & $\underline{39.0}_{0.04}$ / $\underline{34.0}_{0.04}$ / $\underline{35.9}_{0.00}$ \\
    \midrule
    \multicolumn{1}{l}{\textbf{\textit{With Llama-3-8B-Instruct (8.03B)}}} & \multicolumn{6}{l}{} \\
    \midrule
        BM25        & $56.9_{0.61}$ / $56.5_{0.53}$
        & $45.2_{0.37}$ / $36.5_{0.98}$
        & $0.328_{0.01}$ / $0.664_{0.02}$
        & $16.7_{0.33}$ / $15.1_{0.37}$ / $14.7_{0.33}$
        & $41.0_{0.02}$ / $35.2_{0.04}$ / $\underline{40.6}_{0.02}$
        & $31.4_{0.16}$ / $26.6_{0.16}$ / $27.6_{0.24}$ \\
        Contriever
        & $58.5_{0.20}$ / $\underline{58.1}_{0.41}$
        & $47.2_{0.37}$ / $39.1_{0.29}$
        & $0.314_{0.00}$ / $\underline{0.631}_{0.01}$
        & $\underline{17.2}_{0.04}$ / $\underline{15.6}_{0.04}$ / $\underline{15.1}_{0.04}$
        & $41.1_{0.08}$ / $35.4_{0.12}$ / $40.5_{0.04}$
        & $\underline{32.1}_{0.16}$ / $\underline{27.2}_{0.24}$ / $\underline{28.5}_{0.16}$ \\
        IC-RALM
        & $\underline{59.3}_{0.04}$ / $56.6_{0.29}$
        & $38.0_{0.82}$ / $29.9_{0.41}$
        & $0.351_{0.01}$ / $0.671_{0.01}$
        & $14.1_{0.24}$ / $12.5_{0.24}$ / $12.1_{0.08}$
        & $37.5_{0.08}$ / $31.5_{0.04}$ / $39.2_{0.04}$
        & $29.2_{0.69}$ / $24.6_{0.57}$ / $25.6_{0.49}$ \\
        REPLUG
        & $55.5_{1.06}$ / $47.5_{2.00}$
        & $42.0_{1.35}$ / $31.8_{1.14}$
        & $0.314_{0.00}$ / $0.633_{0.00}$
        & $14.8_{0.04}$ / $13.2_{0.00}$ / $11.8_{0.04}$
        & $\textbf{42.5}_{0.08}$ / $\textbf{37.1}_{0.06}$ / $\textbf{40.8}_{0.12}$
        & $30.4_{0.20}$ / $26.1_{0.12}$ / $25.8_{0.37}$ \\
        RankGPT (Llama-3-8B-Instruct)
        & $57.0_{0.20}$ / $56.3_{0.12}$
        & $47.2_{0.90}$ / $38.4_{0.53}$
        & $0.318_{0.01}$ / $0.637_{0.01}$
        & $16.9_{0.20}$ / $15.3_{0.16}$ / $14.8_{0.29}$
        & $41.0_{0.04}$ / $35.4_{0.08}$ / $40.5_{0.16}$
        & $31.0_{0.12}$ / $26.3_{0.08}$ / $27.3_{0.16}$ \\
        RankGPT (GPT-5-nano)
        & \textbf{59.5} / 58.0
        & 45.1 / 36.2
        & 0.321 / 0.638
        & 17.1 / 15.4 / 15.0
        & 41.0 / 35.3 / 40.5
        & 31.5 / 26.5 / 27.8 \\
        ICR (Llama-3-8B-Instruct)
        & $58.4_{0.29}$ / $57.3_{0.37}$
        & $\underline{48.0}_{0.37}$ / $\underline{39.3}_{0.73}$
        & $\underline{0.312}_{0.01}$ / $\underline{0.631}_{0.02}$
        & $17.1_{0.04}$ / $15.4_{0.00}$ / $14.9_{0.00}$
        & $41.3_{0.16}$ / $35.5_{0.29}$ / $\textbf{40.8}_{0.24}$
        & $31.8_{0.33}$ / $26.8_{0.24}$ / $28.1_{0.29}$ \\
        \rowcolor{green!15} PURPLE (Ours)
        & $59.2_{0.82}$ / $\textbf{58.8}_{0.78}$
        & $\textbf{49.6}_{0.65}$ / $\textbf{41.6}_{0.53}$
        & $\textbf{0.307}_{0.01}$ / $\textbf{0.624}_{0.01}$
        & $\textbf{17.6}_{0.04}$ / $\textbf{15.9}_{0.00}$ / $\textbf{15.3}_{0.08}$
        & $\underline{41.4}_{0.49}$ / $\underline{35.8}_{0.73}$ / $40.3_{0.37}$
        & $\textbf{32.5}_{0.04}$ / $\textbf{27.5}_{0.00}$ / $\textbf{28.8}_{0.04}$ \\
    \midrule
    \multicolumn{1}{l}{\textbf{\textit{With Llama-3-70B-Instruct (70.6B)}}} & \multicolumn{6}{l}{} \\
    \midrule
        BM25
        & $71.3_{0.37}$ / $70.8_{0.37}$
        & $54.5_{0.37}$ / $47.2_{0.41}$
        & $0.245_{0.01}$ / $0.544_{0.01}$
        & $18.1_{0.33}$ / $16.5_{0.29}$ / $14.9_{0.37}$
        & $43.8_{0.53}$ / $38.2_{0.41}$ / $40.5_{0.49}$
        & $36.3_{0.16}$ / $30.9_{0.20}$ / $33.0_{0.12}$ \\
        Contriever
        & $71.2_{0.78}$ / $70.8_{0.73}$
        & $56.6_{0.16}$ / $49.5_{0.29}$
        & $\underline{0.238}_{0.00}$ / $0.528_{0.00}$
        & $18.6_{0.08}$ / $\underline{17.0}_{0.08}$ / $\underline{15.6}_{0.08}$
        & $44.1_{0.08}$ / $\underline{38.6}_{0.12}$ / $41.1_{0.00}$
        & $36.5_{0.00}$ / $\underline{31.4}_{0.00}$ / $\underline{33.3}_{0.00}$ \\
        IC-RALM
        & $66.5_{0.04}$ / $66.4_{0.00}$
        & $49.0_{0.29}$ / $41.8_{0.12}$
        & $0.261_{0.00}$ / $0.555_{0.00}$
        & $14.8_{0.02}$ / $13.3_{0.04}$ / $12.2_{0.00}$
        & $39.7_{0.00}$ / $34.1_{0.24}$ / $38.6_{0.16}$
        & $32.0_{0.20}$ / $27.4_{0.00}$ / $28.8_{0.20}$ \\
        REPLUG
        & $65.8_{0.33}$ / $65.4_{0.41}$
        & $51.7_{0.04}$ / $44.0_{0.08}$
        & $0.258_{0.01}$ / $0.554_{0.02}$
        & $15.2_{0.04}$ / $13.9_{0.04}$ / $12.1_{0.04}$
        & $41.8_{0.12}$ / $36.6_{0.33}$ / $38.6_{0.35}$
        & $32.1_{0.08}$ / $27.6_{0.12}$ / $27.7_{0.08}$ \\
        RankGPT (Llama-3-8B-Instruct)
        & $69.7_{0.16}$ / $69.1_{0.12}$
        & $\underline{56.9}_{0.16}$ / $48.8_{0.41}$
        & $0.247_{0.00}$ / $0.546_{0.01}$
        & $18.0_{0.24}$ / $16.5_{0.29}$ / $15.2_{0.29}$
        & $44.0_{0.04}$ / $\underline{38.6}_{0.08}$ / $41.1_{0.12}$
        & $35.8_{0.00}$ / $30.6_{0.00}$ / $32.4_{0.08}$ \\
        RankGPT (GPT-5-nano)
        & \textbf{73.8} / \textbf{73.5}
        & 55.3 / 48.2
        & 0.240 / \underline{0.523}
        & \underline{18.7} / \underline{17.0} / \textbf{15.8}
        & \textbf{44.6} / \textbf{38.8} / \textbf{41.5}
        & \underline{36.6} / 31.3 / 33.2 \\
        ICR (Llama-3-8B-Instruct)
        & $71.6_{0.16}$ / $71.0_{0.16}$
        & $56.8_{0.24}$ / $\underline{49.6}_{0.57}$
        & $\underline{0.238}_{0.00}$ / $0.531_{0.00}$
        & $18.3_{0.00}$ / $16.8_{0.04}$ / $15.3_{0.16}$
        & $44.3_{0.06}$ / $38.5_{0.04}$ / $\underline{41.2}_{0.20}$
        & $36.3_{0.16}$ / $31.1_{0.24}$ / $33.2_{0.20}$ \\
        \rowcolor{green!15} PURPLE (Ours)
        & $\underline{72.2}_{0.49}$ / $\underline{71.8}_{0.57}$
        & $\textbf{57.0}_{0.08}$ / $\textbf{49.9}_{0.41}$
        & $\textbf{0.236}_{0.02}$ / $\textbf{0.520}_{0.01}$
        & $\textbf{18.8}_{0.04}$ / $\textbf{17.1}_{0.00}$ / $\underline{15.6}_{0.08}$
        & $\underline{44.4}_{0.02}$ / $\textbf{38.8}_{0.06}$ / $41.0_{0.04}$
        & $\textbf{37.3}_{0.01}$ / $\textbf{32.1}_{0.04}$ / $\textbf{34.0}_{0.06}$ \\
        \midrule
    \end{tabular}}
    \label{tab:lamp}
\end{table*}

%% file: sections/05-results.tex
\begin{figure}[t]
    \centering
    \includegraphics[width=0.8\linewidth]{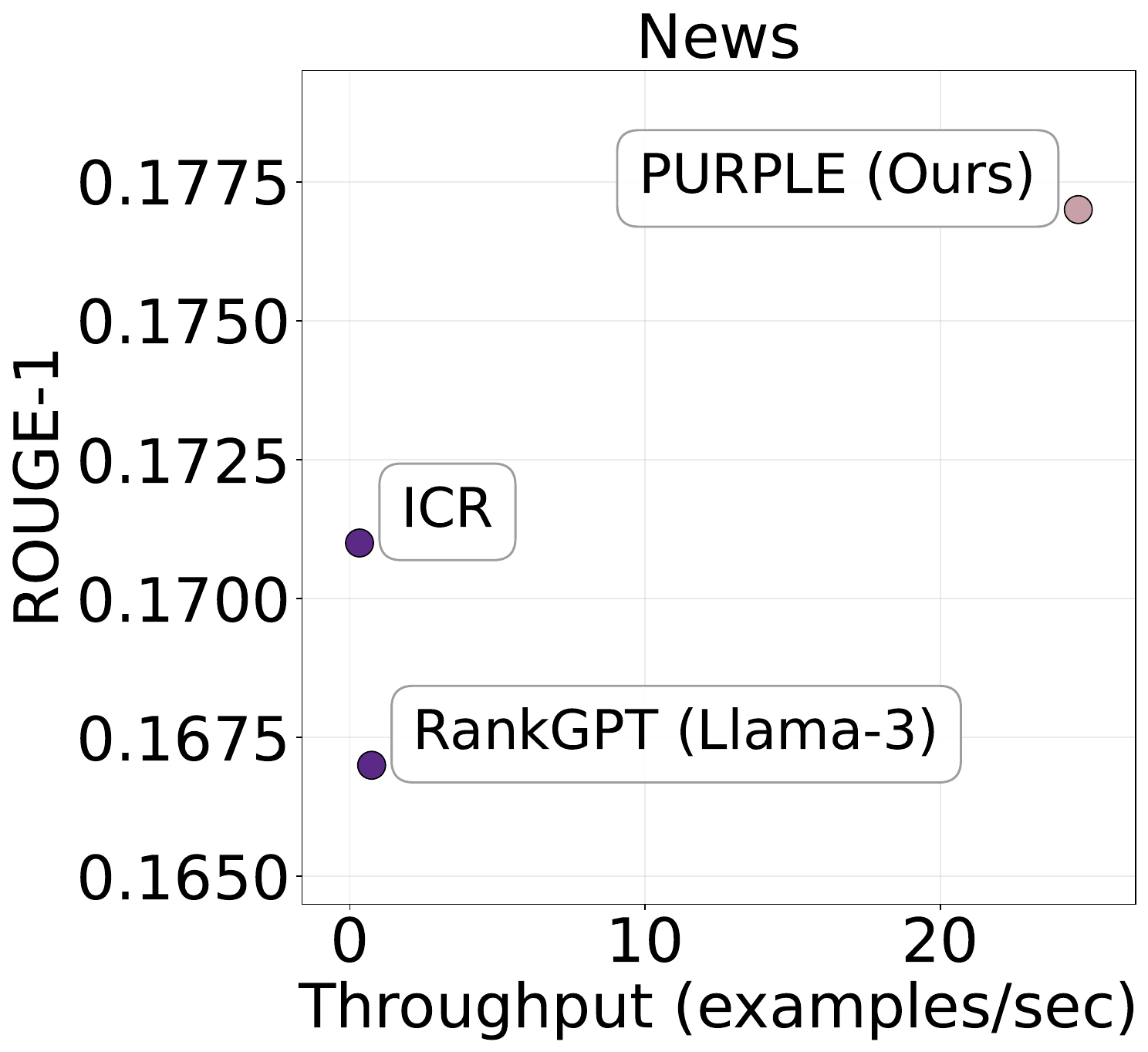}
    \caption{Performance--throughput tradeoff on News.}
    \label{fig:tradeoff_news}
\end{figure}

\section{Results and Analysis}
\label{sec:results}

\subsection{Overall Performance Comparison}

Table~\ref{tab:lamp} presents the results of PURPLE and the baselines on the LaMP benchmark.
Table~\ref{tab:longlamp} in the Appendix presents the results on the LongLaMP benchmark.
The main findings are as follows:

\paragraph{PURPLE consistently outperforms strong baselines across LLM scales.}

Across all tasks and LLMs of varying sizes, PURPLE achieves consistent improvements over existing methods.
Compared with Contriever, which is of comparable model size, PURPLE's propensity scores provide more effective ranking signals than raw relevance.
Compared with zero-shot rerankers, which use much larger backbone LLMs and incur higher inference cost, PURPLE achieves stronger personalization with a much smaller model, since reinforcement learning training help better capture the utility of multiple records.
Compared with in-context RALMs, which provide user records one at a time to the LLM and combine multiple records post hoc, our single-stage modeling more effectively captures personalized signals, highlighting the advantage of treating user profiles holistically.

\paragraph{PURPLE outperforms strong baselines with high computational throughput.}

Figure~\ref{fig:tradeoff_news} shows that, on a representative LaMP dataset, PURPLE outperforms existing methods while being more efficient.
Specifically, PURPLE achieves higher performance than ICR and RankGPT (Llama-3) while maintaining high computational throughput.
Additional results across the six LaMP tasks are provided in Figure~\ref{fig:tradeoff_all} in the Appendix.

\paragraph{PURPLE is effective across task types, including regression.}

Since our reward is based on the log-likelihood that the LLM assigns to the reference response, it does not directly reflect numerical distances between regression targets.
Nevertheless, PURPLE still achieves strong gains on the regression task (Rating).
This demonstrates that log-likelihood provides a principled and broadly applicable reward signal across diverse task forms.

\input{tables/ablation}

\subsection{Ablation Studies on Model Architecture and Reward Modeling}
\label{sec:ablation}

Table~\ref{tab:ablation} presents the ablation results of PURPLE using \texttt{Phi-4-Mini-Instruct}.
Overall, we examine three key design choices.
First, instead of performing token-level cross-attention, we test a simplified variant that pools the query into a single embedding vector, which is then prepended as an extra token to the Transformer encoder (w/o CA).
This approach is less effective, indicating that fine-grained token-level interactions between the query and the user records are crucial for accurate personalization.
Second, we remove the Transformer encoder entirely, resulting in a point-wise scoring model where each record is processed independently (w/o RDM).
This variant shows the largest performance drop across tasks.
Although the model can still leverage individually informative records, it fails to model dependencies such as redundancy and complementarity among records.
Third, we examine an alternative reward design that uses task-specific evaluation metrics as the reward (Accuracy for classification, MAE for regression, and ROUGE-1 for generation).
This metric-based reward is substantially more coarse-grained, providing weaker learning signals and making optimization more challenging.
This variant consistently underperforms PURPLE across all tasks, thereby strongly motivating our log-likelihood-based reward.

These results highlight that token-level cross-attention, inter-record dependency modeling, and the log-likelihood-based reward are all indispensable, validating our design choices of treating user profiles as structured contexts rather than isolated records, and providing semantically richer reward signals via reference log-likelihoods.

\subsection{Analysis on TopK Selection at Inference}
\label{sec:topk}

\begin{figure}[t]
    \centering
    \includegraphics[width=\linewidth]{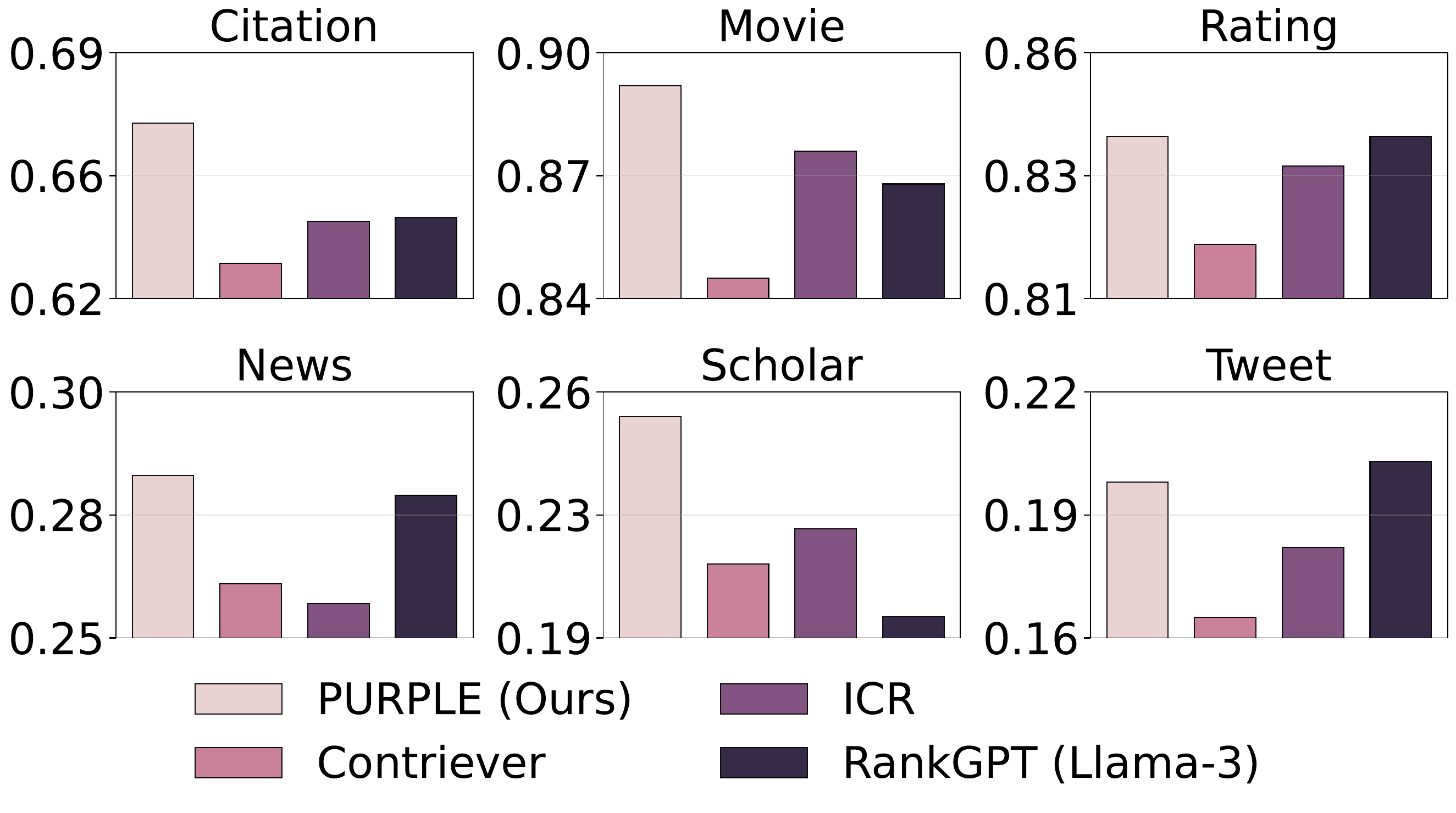}
    \caption{Ranking accuracy across the six LaMP tasks.}
    \label{fig:ranking}
\end{figure}

To further justify our inference strategy of selecting the top-$K$ records based on the propensity scores, we compare the top-$K$ selections of PURPLE, ICR, RankGPT, and Contriever on the six LaMP datasets.
For each test set example, we consider the top-5 records proposed by each method and randomly sample five orderings from the $5! = 120$ permutations as controls.
These six orderings are ranked based on downstream generation quality using \texttt{Llama-3-8B-Instruct}, as measured by task-specific evaluation metrics.
As shown in Figure~\ref{fig:ranking}, orderings induced by PURPLE's propensity scores are more frequently ranked as the best among the six candidates.
This indicates that PURPLE's scoring function better captures relative dependencies between records, highlighting that our method not only identifies useful records but also arranges them in an order that maximizes downstream personalization utility.
We attribute this advantage of PURPLE to its holistic modeling of the user profile, and emphasize that our method adopts a profile-level optimization strategy rather than relying solely on local, pairwise relevance.

\subsection{Human Evaluation and Case Study}
\label{sec:human}

To complement our automatic evaluation and verify the practical utility of the selected profiles, we conduct a rigorous side-by-side human evaluation.
While automatic metrics measure lexical overlap, they often fail to capture the subtle stylistic nuances, such as voice, stance, and informal phrasing, that define true personalization.
We focus our human evaluation on \textit{Personalized Tweet Paraphrasing} (Tweet), which requires significant stylistic adaptation to user personas.
In this study, we randomly sample $250$ instances.
For each instance, human evaluators are presented with the input query, a snippet of the user's history, and two anonymized model outputs: one generated using the recent baseline, ICR, and the other using PURPLE.
Evaluators perform a blind, forced-choice comparison to determine which output better reflects the user's specific persona while maintaining semantic fidelity.
The results demonstrate a strong evaluator preference for PURPLE, surpassing ICR by an absolute margin of $14.4\%$ ($57.2\%$ vs. $42.8\%$).
This confirms that our method effectively captures personal stylistic features, such as slang usage and informal tone, where heuristic relevance retrieval often fails. 
We provide detailed qualitative case studies analyzing these behaviors in Appendix~\ref{app:case}.

\begin{figure*}
    \centering
    \includegraphics[width=\linewidth]{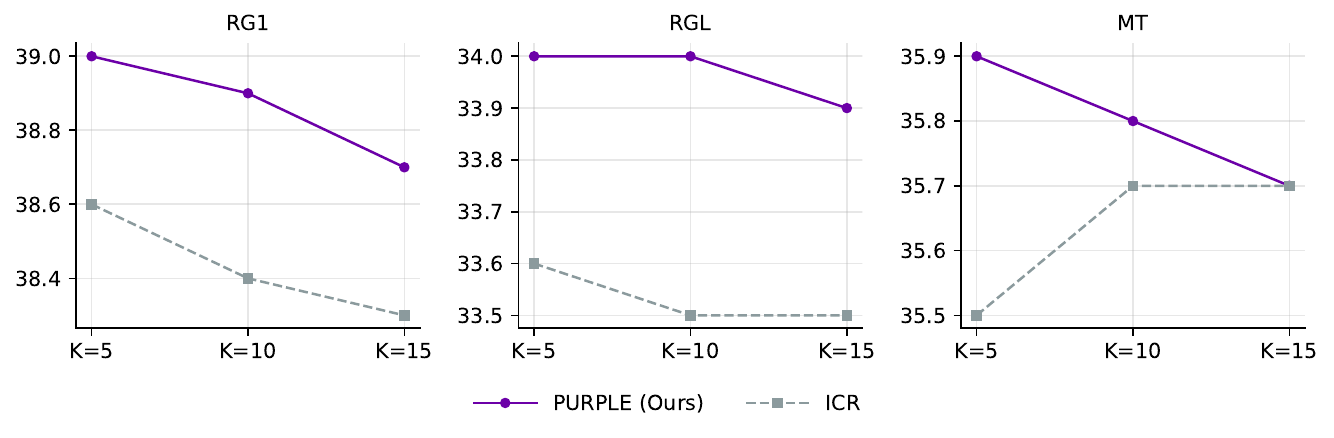}
    \caption{Effect of varying profile sizes on Tweet.}
    \label{fig:profile}
\end{figure*}

\subsection{Sensitivity to Profile Size}

While the main experiments employ a profile size of $K = 5$, we further investigate the impact of varying profile sizes $K \in \{5, 10, 15\}$ on \textit{Personalized Tweet Paraphrasing} (Tweet) using \texttt{Phi-4-Mini-Instruct}.
Throughout this analysis, the size of the candidate pool remains fixed at 20.
As illustrated in Figure~\ref{fig:profile}, PURPLE consistently outperforms ICR across all settings.
However, we observe that increasing $K$ beyond 5 generally leads to a slight performance degradation.
This finding aligns with prior results in the RAG literature, which suggest that larger contexts may introduce redundancy or interference rather than additional utility~\citep{yu2024rankrag,salemi2024lamp}.

%% file: tables/ablation.tex
\begin{table*}[t]
    \centering
    \caption{Ablation studies of PURPLE using \texttt{Phi-4-Mini-Instruct}. CA and RDM stand for cross-attention and inter-record dependency modeling, respectively. MR refers to using task-specific metrics as the reward.}
    \vspace{1mm}
    \adjustbox{max width=\linewidth}{
    \begin{tabular}{l|cc|c|ccc}
        \toprule
        \textbf{Task}
            & \textbf{Citation}
            & \textbf{Movie}
            & \textbf{Rating}
            & \textbf{News}
            & \textbf{Scholar}
            & \textbf{Tweet} \\
        \midrule
        \textbf{Metric}
            & Acc. / F1
            & Acc. / F1
            & MAE / RMSE
            & RG1 / RGL / MT
            & RG1 / RGL / MT
            & RG1 / RGL / MT \\
        \midrule
        PURPLE
            & \textbf{66.2} / \textbf{65.8}
            & \textbf{38.2} / \textbf{33.6}
            & \textbf{0.405} / \textbf{0.788}
            & \textbf{15.2} / \textbf{13.5} / \textbf{12.5}
            & \textbf{40.0} / \textbf{33.5} / \textbf{42.4}
            & \textbf{39.1} / 34.0 / 35.9 \\
        \hspace{5mm}w/o CA
            & 64.8 / 64.5
            & 35.1 / 29.7
            & 0.440 / 0.816
            & 14.8 / 13.2 / 12.4
            & \textbf{40.0} / \textbf{33.5} / 42.2
            & \textbf{39.1} / \textbf{34.1} / 36.0 \\
        \hspace{5mm}w/o RDM
            & 61.3 / 60.6
            & 35.0 / 31.1
            & 0.449 / 0.850
            & 14.5 / 12.8 / 11.9
            & 39.7 / 33.1 / 41.9
            & 39.0 / 34.0 / \textbf{36.1} \\
        \midrule
        \hspace{5mm}w/ MR
            & 64.8 / 64.8
            & 38.0 / 33.3
            & 0.433 / 0.854
            & 15.0 / 13.2 / 12.4
            & 39.5 / 32.9 / 41.8
            & 38.7 / 33.7 / 35.6 \\
        \midrule
    \end{tabular}}
    \label{tab:ablation}
\end{table*}

%% file: sections/06-conclusion.tex
\section{Conclusion}
\label{sec:conclusion}

In this work, we study the problem of retrieval-augmented personalization for LLMs.
Our key intuitions are twofold: (i) \textit{record relevance does not reliably predict personalization utility}, and (ii) \textit{utility is non-monotonic across records}, making greedy aggregation suboptimal.
To address these limitations, we propose PURPLE, a contextual bandit framework that optimizes user profiles by leveraging downstream performance as feedback.
PURPLE jointly models query--record interactions and inter-record dependencies, enabling adaptive selection of user profiles beyond static heuristics.
Instead of task-specific metrics, PURPLE relies on the log-likelihood that the LLM assigns to the reference response as the reward, providing a semantically richer reward signal that facilitates more effective training.
Extensive experiments on nine real-world personalization tasks spanning classification, regression, and generation show that PURPLE consistently outperforms heuristic retrievers, LLM-based rerankers, and in-context RALMs, while being more efficient.
These results establish contextual bandit-based retrieval as a principled and scalable paradigm for personalizing LLMs.

%% file: sections/07-limitations.tex
\section*{Limitations}

While PURPLE demonstrates strong effectiveness across personalization tasks, it has several limitations.
First, the method relies on the log-likelihood of reference personalized responses under a frozen LLM as the training reward, which assumes access to high-quality personalized supervision.
In practical deployment, such explicit supervision may be sparse or unavailable, as user feedback often takes implicit or noisy forms (e.g., engagement signals).
Extending PURPLE to effectively leverage these weaker signals remains an open challenge.
Second, although using response log-likelihood as the reward provides a unified optimization signal, PURPLE is still trained separately for each task in our current study.
We do not explicitly evaluate the extent to which a policy learned under this unified objective can generalize to unseen tasks or domains.
Exploring more transferable training and deployment paradigms is thus left for future work.

%% file: sections/99-appendix.tex
\appendix

\input{tables/longlamp}

\section{Details of Gradient Estimation}
\label{app:gradient}

To estimate the gradient in Equation~\ref{eq:gradient}, we first draw a batch of examples $\{(\mathcal{H}_b, x_b, y_b)\}_{b=1}^{B}$.
For each example, we sample $M$ profiles $\mathcal{P}_b^1, \dots, \mathcal{P}_b^M$ from $\pi_\theta(\cdot \mid \mathcal{C}_b)$, and compute the empirical mean of the gradient estimates.
This learning procedure corresponds to the REINFORCE algorithm~\citep{sutton1999policy}, with the Monte Carlo gradient estimate:
\begin{equation}
\begin{aligned}
    \nabla_{\theta} \mathcal{J}&(\theta) \approx \\
    &\frac{1}{B}\sum_{b=1}^{B} \frac{1}{M} \sum_{m=1}^{M} \nabla_{\theta} \log \pi_\theta(\mathcal{P}_b^m \mid \mathcal{C}_b) \tilde{r}_b^m.
\end{aligned}
\label{eq:monte_carlo}
\end{equation}
To reduce variance in gradient estimation, we apply reward normalization over the $M$ profiles sampled for each example.
Concretely, for each example with rewards $\vr_b = [r_b^1, \dots, r_b^M]^\top$, where $r_b^m = R(\phi(\mathcal{P}_b^m, x_b), y_b)$, the normalized reward is computed as $\tilde{r}_b^m = \frac{r_b^m - \mathrm{mean}(\vr_b)}{\mathrm{std}(\vr_b)}$.

\section{Motivating our Reward}
\label{app:reward}

The choice of using response log-likelihood as the reward is grounded in the generative modeling formulation of RAG~\citep{lewis2020retrievalaugmented}, where the user profile is treated as a latent variable and the response likelihood is obtained by marginalizing over all possible profiles.
Applying Jensen’s inequality to the maximum likelihood objective gives:
\begin{equation}
\begin{aligned}
    &\E_{(\mathcal{H}, x, y) \sim \mathcal{D}} \\
    & \!\left[ \log \sum_{\mathcal{P} \in \mathrm{Perm}_K(\mathcal{H})} \pi_\theta(\mathcal{P} \mid \mathcal{C}) p_\phi(y \mid \mathcal{P}, x) \right] \\
    &\ge \E_{(\mathcal{H}, x, y) \sim \mathcal{D}, \mathcal{P} \sim \pi_\theta(\cdot \mid \mathcal{C})}[\log p_\phi(y \mid \mathcal{P}, x)].
\end{aligned}
\end{equation}
Therefore, maximizing the expected log-likelihood is equivalent to optimizing the evidence lower bound (ELBO), with $p_\phi$ modeled by a frozen LLM.

\section{Discussion of Alternative Personalization Methods}
\label{app:alternative}

This section provides a brief discussion of personalization methods that fall outside our experimental setting.
These approaches are not included in our empirical comparisons because they operate under different assumptions or training regimes, which prevents a fair and controlled evaluation against our method.
We review them here to contextualize our design choices and to clarify the scope of the methods considered in this work.

In their benchmark paper, \citet{salemi2024lamp} propose two baseline LLM personalization strategies: In-Prompt Augmentation (IPA) and Fusion-in-Decoder (FiD).
IPA represents the standard processing applied in our method as well as in all compared baselines: it integrates all retrieved records into a single prompt and feeds them to a frozen LLM.
FiD, on the other hand, encodes multiple inputs separately within the encoder, producing representations that are attended to by the decoder; as a result, it relies on encoder-decoder architectures and requires fine-tuning of the LLM.

PREMIUM~\cite{sun2025premium} utilizes a pre-defined tag library to construct candidate tag sets from user queries using a neural network.
The network is trained by leveraging users’ preference feedback on LLM responses generated from different candidate tag sets.
While a suitable tag set can be readily identified for certain personalization tasks, such as \textit{Personalized Movie Tagging} (Movie), this is generally not the case for open-ended personalized generation.
Moreover, the tag sets considered in their work correspond to static topical features (e.g., Investment, Bakery, Comedy), whereas personalization may involve more nuanced attributes such as writing style, tone, and language use.
These attributes are difficult to capture with a fixed set of tags but are more naturally reflected in user history records.
We consider a more general setting in which only a collection of history records (and no online user feedback) is available, and we aim to select a useful subset for retrieval-augmented personalized generation.

Personalized Pieces~\cite{tan2024personalized} trains LoRA adapters for representative users and combines them for new users based on user profile similarity, while \citet{li2024personalized} incorporates user information by fine-tuning the LLM under a personalized training objective.
As a result, these works fall outside the class of RAG-based methods considered in our comparisons.

\begin{figure*}[t]
    \centering
    \includegraphics[width=\linewidth]{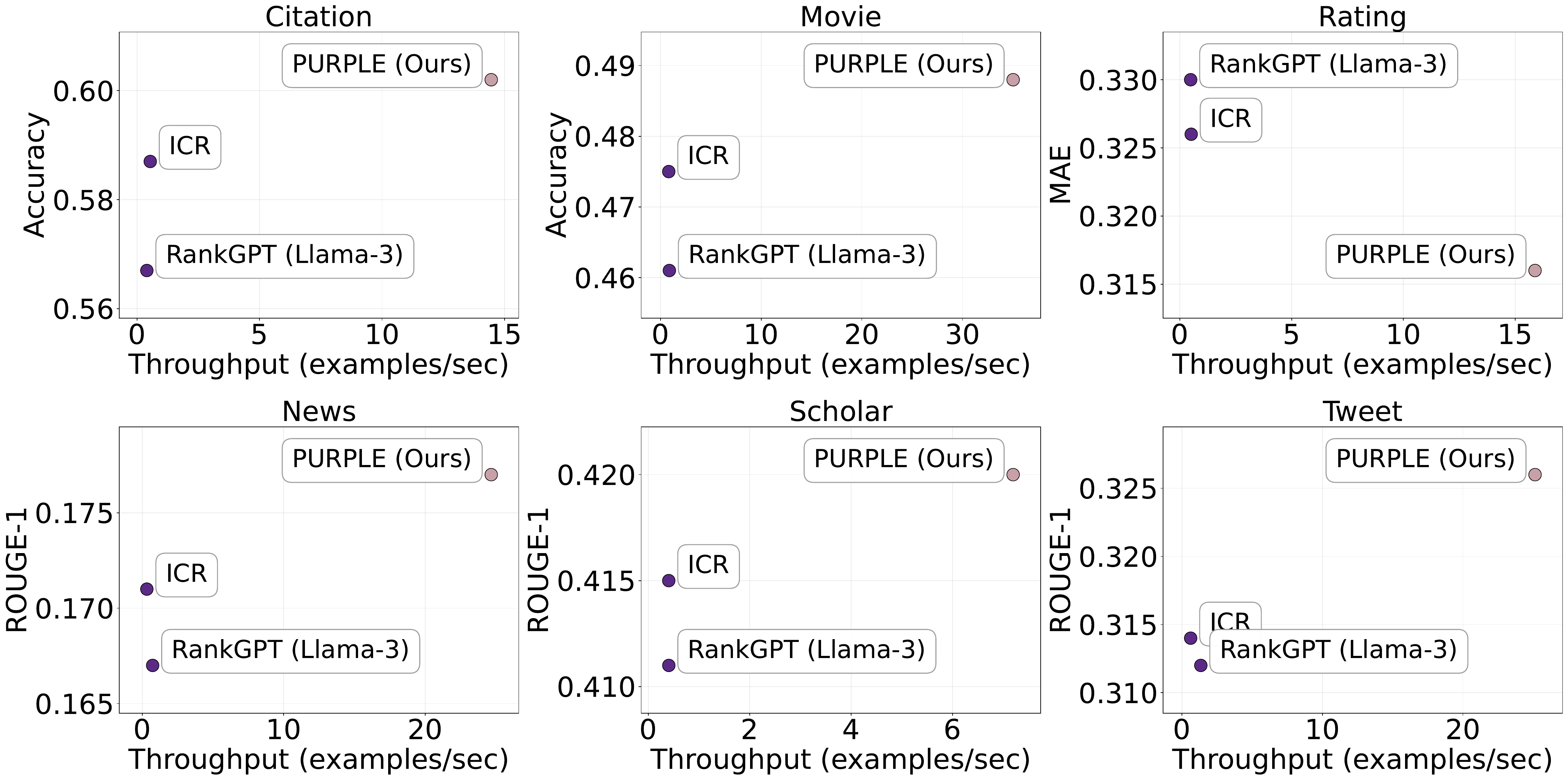}
    \caption{Performance--throughput tradeoff across the six LaMP tasks.}
    \label{fig:tradeoff_all}
\end{figure*}

\section{Implementation Details}
\label{app:implementation}

We employ a frozen pre-trained Contriever to first encode both queries and user history records into token embeddings.
The only trainable components are the remaining modules of the user records encoder.
These include a cross-attention layer that integrates query information into record embeddings, a Transformer encoder that captures inter-record dependencies, and an MLP decoder that maps the updated record encodings into scalar propensity scores.
We set the number of Transformer encoder layers to $l = 12$, resulting in a parameter size roughly twice that of Contriever, while still being substantially smaller than the zero-shot rerankers and in-context RALMs used as baselines.
For gradient estimation, we use a batch size of $B = 16$ and sample $M = 32$ user profiles for each example.
We train the model for 10 epochs using the Adam optimizer~\citep{kingma2015adam} with $\beta_{1} = 0.9$, $\beta_{2} = 0.999$, and a learning rate of $1 \times 10^{-4}$.
During training, we apply a gradient clipping norm of 1.0.
The checkpoint achieving the best validation performance is selected for testing.

In all experiments, we use frozen LLMs both to generate personalized responses and to evaluate the log-likelihood of reference responses conditioned on the query and user profiles (i.e., our reward).
For generation, we set the temperature to $T = 0.7$ and employ nucleus sampling~\citep{holtzman2020curious} with $top\_p = 0.8$.
For \texttt{Phi-4-Mini-Instruct} and \texttt{Llama-3-8B-Instruct}, we deploy them on a single NVIDIA H100 GPU.
For \texttt{Llama-3-70B-Instruct}, we deploy the model across four NVIDIA H100 GPUs using vLLM~\citep{kwon2023efficient}.
All LLMs are deployed in BF16 precision.

\section{Qualitative Case Studies}
\label{app:case}

In this section, we present a detailed case study from our human evaluation where PURPLE and the baseline ICR diverge significantly, illustrating the qualitative superiority of our method.
In the Tweet task, preserving the logic of the original sentence is as critical as matching the user's informal voice.
The example below demonstrates a failure in logical reasoning by ICR.
The input tweet contains a comparative structure (``at least we attended... unlike someone'').
The baseline fails to parse this dependency, resulting in a paraphrase that states the exact opposite of the truth.
PURPLE successfully disentangles the complex sentence structure and preserves the correct meaning.

\vspace{1em}
\noindent\fbox{
    \parbox{0.95\linewidth}{
        \small
        \textbf{User Persona / History:}
        \begin{itemize}[leftmargin=*, noitemsep, topsep=0pt]
            \item @tommoger thanks tom, and glad it all went well
            \item @callamKAPOW I tried to change it but the pic was to big and I can't find the old one
            \item Thinks I can pass English, I just wish I could be as confident about geography
        \end{itemize}

        \vspace{1em}
        \hrule
        \vspace{1em}

        \textbf{Input Tweet:} \\
        Paraphrase the following tweet without any explanation before or after it: @picklesticks6 and I am in the same boat, but at least we attended the math lessons unlike someone we both know who is taking their exam today and feels sad about it. =(

        \vspace{0.5em}
        \textbf{Reference Response:} \\
        @picklesticks6 same, but at least we turned up to the maths lessons unlike someone we know whos takeing their exam today =(

        \vspace{1em}
        \hrule
        \vspace{1em}

        \textbf{ICR (Baseline) Generation:} \\
        @picklesticks6 we both missed math classes, and it's making our exam prep tougher, unlike someone who went and feels down about it today. =(

        \vspace{0.5em}
        \textbf{PURPLE (Ours) Generation:} \\
        @picklesticks6 feels the same about math exams as me, but at least we went to the lessons, unlike someone else who feels sad today about their exam. =(

        \vspace{1em}
        \hrule
        \vspace{1em}

        \textbf{Analysis:} \\
        This case highlights a critical semantic failure in the baseline. \textbf{ICR} generates ``we both missed math classes,'' which directly contradicts the input stating ``at least we attended.'' It effectively swaps the roles of the speaker and the third party. \textbf{PURPLE} demonstrates superior natural language understanding by correctly preserving the fact that the speaker \textit{did} attend the lesson. Furthermore, PURPLE better matches the user's persona (which discusses academic confidence in English/Geography) by maintaining the conversational flow without losing the factual premise.
    }
}

\section{LLM Usage}

During the preparation of this manuscript, we employed LLMs to assist with English writing refinement and style polishing.
All technical content, including the design of PURPLE, theoretical formulations, experimental setup, and reported results, was conceived, implemented, and validated by the authors.
The LLMs were used solely for linguistic improvement and did not contribute to the research methodology or experimental findings.

%% file: tables/longlamp.tex
\begin{table*}[t]
    \centering
    \caption{Results on the LongLaMP benchmark. The best and second-best results in each column are highlighted in \textbf{bold} and \underline{underlined}, respectively. We report the mean and standard deviation over three runs with different random seeds for LLM generation. For \texttt{GPT-5-nano}, we report only a single run due to API cost constraints.}
    \adjustbox{max width=\linewidth}{
    \begin{tabular}{l|ccc}
        \toprule
        \textbf{Task}
            & \textbf{Abstract}
            & \textbf{Topic}
            & \textbf{Review} \\
        \midrule
        \textbf{Metric}
            & R1 / RL / MT
            & R1 / RL / MT
            & R1 / RL / MT \\
    \midrule
    \multicolumn{1}{l}{\textbf{\textit{With Phi-4-Mini-Instruct (3.84B)}}} & \multicolumn{3}{l}{} \\
    \midrule
        BM25
        & $38.6_{0.12}$ / $22.0_{0.16}$ / $26.2_{0.08}$
        & $25.2_{0.41}$ / $12.7_{0.20}$ / $\underline{17.1}_{0.16}$
        & $25.8_{1.39}$ / $13.2_{0.53}$ / $16.5_{0.16}$ \\
        Contriever
        & $38.5_{0.04}$ / $21.9_{0.12}$ / $26.1_{0.08}$
        & $\underline{25.7}_{1.80}$ / $\underline{13.0}_{0.73}$ / $16.6_{0.20}$
        & $25.2_{1.92}$ / $13.1_{0.57}$ / $15.9_{0.78}$ \\
        IC-RALM
       & $37.2_{0.04}$ / $21.1_{0.02}$ / $25.1_{0.04}$
        & $24.8_{1.47}$ / $12.4_{0.78}$ / $15.9_{0.20}$
        & $25.0_{1.35}$ / $12.7_{0.49}$ / $15.8_{0.16}$ \\
        REPLUG
        & $36.0_{0.24}$ / $21.2_{0.20}$ / $23.8_{0.20}$
        & $23.5_{1.06}$ / $12.1_{0.20}$ / $13.1_{0.41}$
        & $20.8_{2.94}$ / $11.2_{1.14}$ / $12.5_{1.63}$ \\
        RankGPT (Llama-3-8B-Instruct)
        & $38.6_{0.12}$ / $22.0_{0.08}$ / $26.2_{0.08}$
        & $25.1_{0.53}$ / $12.7_{0.29}$ / $\underline{17.1}_{0.18}$
        & $25.1_{1.63}$ / $12.9_{0.53}$ / $15.8_{0.45}$ \\
        RankGPT (GPT-5-nano)
        & \textbf{39.1} / \textbf{22.4} / \textbf{26.9}
        & 24.9 / 12.5 / \textbf{17.5}
        & \underline{27.1} / \underline{13.7} / \underline{16.6} \\
        ICR (Llama-3-8B-Instruct)
        & $38.5_{0.24}$ / $22.0_{0.16}$ / $26.2_{0.16}$
        & $\underline{25.7}_{1.71}$ / $\underline{13.0}_{0.73}$ / $16.5_{0.24}$
        & $25.8_{1.67}$ / $13.2_{0.53}$ / $16.4_{0.45}$ \\
        \rowcolor{green!15} PURPLE (Ours)
        & $\underline{38.8}_{0.08}$ / $\underline{22.1}_{0.16}$ / $\underline{26.4}_{0.12}$
        & $\textbf{26.2}_{1.14}$ / $\textbf{13.2}_{0.61}$ / $\textbf{17.5}_{0.12}$
        & $\textbf{27.9}_{0.04}$ / $\textbf{14.4}_{0.08}$ / $\textbf{17.1}_{0.04}$ \\
    \midrule
    \multicolumn{1}{l}{\textbf{\textit{With Llama-3-8B-Instruct (8.03B)}}} & \multicolumn{3}{l}{} \\
    \midrule
        BM25
        & $42.2_{0.04}$ / $\underline{24.2}_{0.04}$ / $31.8_{0.08}$
        & $29.8_{0.73}$ / $14.6_{0.24}$ / $\underline{20.2}_{0.12}$
        & $\underline{30.9}_{2.00}$ / $\underline{15.3}_{0.82}$ / $\underline{20.8}_{0.41}$ \\
        Contriever
        & $\underline{42.3}_{0.24}$ / $\underline{24.2}_{0.29}$ / $31.7_{0.24}$
        & $\underline{30.6}_{1.39}$ / $\textbf{15.2}_{0.53}$ / $20.0_{0.32}$
        & $30.1_{2.49}$ / $14.9_{1.02}$ / $19.7_{0.90}$ \\
        IC-RALM
        & $38.3_{0.90}$ / $20.4_{0.78}$ / $29.1_{0.37}$
        & $25.0_{0.90}$ / $11.4_{1.06}$ / $15.5_{1.92}$
        & $29.8_{1.27}$ / $13.6_{0.98}$ / $18.3_{0.98}$ \\
        REPLUG
        & $38.2_{0.45}$ / $20.7_{0.33}$ / $28.4_{0.20}$
        & $20.2_{1.22}$ / $10.8_{0.57}$ / $11.9_{1.27}$
        & $15.5_{2.04}$ / $8.8_{0.94}$ / $8.8_{1.02}$ \\
        RankGPT (Llama-3-8B-Instruct)
        & $42.2_{0.04}$ / $\underline{24.2}_{0.04}$ / $31.9_{0.04}$
        & $\textbf{30.8}_{1.39}$ / $\textbf{15.2}_{0.57}$ / $\textbf{20.4}_{0.36}$
        & $30.5_{2.45}$ / $15.1_{1.02}$ / $20.4_{0.82}$ \\
        RankGPT (GPT-5-nano)
        & \textbf{42.5} / \textbf{24.5} / \textbf{32.1}
        & 28.7 / 14.2 / \underline{20.2}
        & \textbf{33.6} / \textbf{16.5} / \textbf{21.4} \\
        ICR (Llama-3-8B-Instruct)
        & $41.9_{0.24}$ / $23.9_{0.20}$ / $31.5_{0.16}$
        & $\underline{30.6}_{1.35}$ / $\underline{15.1}_{0.57}$ / $\underline{20.2}_{0.20}$
        & $30.7_{1.96}$ / $15.2_{0.78}$ / $20.3_{0.41}$ \\
        \rowcolor{green!15} PURPLE (Ours)
        & $\underline{42.3}_{0.08}$ / $\underline{24.2}_{0.12}$ / $\underline{32.0}_{0.24}$
        & $30.4_{1.63}$ / $15.0_{0.73}$ / $19.8_{0.24}$
        & $30.2_{1.61}$ / $14.9_{0.83}$ / $20.0_{0.94}$ \\
        \midrule
    \end{tabular}}
    \label{tab:longlamp}
\end{table*}